\pdfoutput=1

\documentclass[11pt]{article}

\usepackage[]{emnlp2021}
\usepackage{booktabs}
\usepackage{times}
\usepackage{latexsym}
\usepackage{multirow}

\usepackage{microtype}
\usepackage{times}
\usepackage{latexsym}

\usepackage{adjustbox}
\usepackage{graphicx}
\usepackage{booktabs}
\usepackage{caption}
\usepackage{subcaption}
\usepackage{marvosym}
\usepackage{mathtools}
\usepackage{multirow}
\usepackage{amsmath}
\usepackage{amssymb}
\usepackage{comment}
\usepackage{adjustbox}
\usepackage{lipsum}
\usepackage{xspace}
\usepackage{makecell}
\usepackage{color}
\usepackage{arydshln}

\usepackage[T1]{fontenc}

\usepackage[utf8]{inputenc}

\usepackage{microtype}

\iffalse
    \newcommand{\an}[1]{\textcolor{cyan}{\bf\small [AN: #1]}}
    \newcommand{\ty}[1]{\textcolor{blue}{\bf\small [TY: #1]}}
    \newcommand{\rui}[1]{\textbf{\textcolor{blue}{[Rui: #1]}}}
    
\else
    \newcommand{\an}[1]{}
    \newcommand{\ty}[1]{}
    \newcommand{\rui}[1]{}
    
\fi

\newcommand{\ie}{\textit{i.e., \xspace}}

\newcommand{\green}[1]{\textcolor{brown}{#1}}
\newcommand{\red}[1]{\textcolor{red}{#1}}
\newcommand{\gray}[1]{\textcolor{gray}{#1}}

\newcommand\blfootnote[1]{%
\begingroup
\renewcommand\thefootnote{}\footnote{#1}%
\addtocounter{footnote}{-1}%
\endgroup
}
\title{An Exploratory Study on Long Dialogue Summarization: \\ What Works and What's Next}

\author{Yusen Zhang$^{*\clubsuit}$ 
\quad Ansong Ni$^{*\dagger}$ 
\quad Tao Yu$^\dagger$ 
\quad Rui Zhang$^\clubsuit$ \\
\bf \quad Chenguang Zhu$^\diamondsuit$ 
\bf \quad Budhaditya Deb$^\diamondsuit$ 
\bf \quad Asli Celikyilmaz$^\ddagger$ \\
\bf \quad Ahmed Hassan Awadallah $^\diamondsuit$ 
\bf \quad Dragomir Radev$^\dagger$ \\
  $^\clubsuit$Penn State University \quad
  $^\dagger$Yale University \quad
  $^\diamondsuit$Microsoft Research \quad
  $^\ddagger$ Facebook AI Research \\
    \texttt{\{yfz5488, rmz5227\}@psu.edu}   \quad    \texttt{asli.ca@live.com} \\
  \texttt{\{ansong.ni, tao.yu, dragomir.radev\}@yale.edu} \\
  \texttt{\{chezhu, Budha.Deb, hassanam\}@microsoft.com } \\
  }

\begin{document}
\maketitle
\blfootnote{$^*$Equal Contribution.}
\blfootnote{$^\ddagger$The work was done when Asli was at MSR.} 

\begin{abstract}
Dialogue summarization helps readers capture salient information from long conversations in meetings, interviews, and TV series.
However, real-world dialogues pose a great challenge to current summarization models, as the dialogue length typically exceeds the input limits imposed by recent transformer-based pretrained models, and the interactive nature of dialogues makes relevant information more context-dependent and sparsely distributed than news articles.
In this work, we perform a comprehensive study on long dialogue summarization by investigating three strategies to deal with the lengthy input problem and locate relevant information: (1) extended transformer models such as Longformer, (2) \textit{retrieve-then-summarize} pipeline models with several dialogue utterance retrieval methods, and (3) hierarchical dialogue encoding models such as HMNet. 
Our experimental results on three long dialogue datasets (QMSum, MediaSum, SummScreen) show that the retrieve-then-summarize pipeline models yield the best performance.
We also demonstrate that the summary quality can be further improved with a stronger retrieval model and pretraining on proper external summarization datasets.
\rui{Add a footnote indicating first two authors contribute equally; Ask Asli about her affiliation}\an{create a github repo and put it in the footnote}
\end{abstract}


\section{Introduction}
Large amount of dialogue data have been produced in meetings, TV series, and interviews~\cite{chen2021summscreen, zhong2021qmsum,zhu2021mediasum}. 
Dialogue summarization aims to generate a short summary for long dialogues to help the readers capture important information more efficiently.

A number of existing works on dialogue summarization focus on extracting the main events of a short conversation \cite{gliwa-etal-2019-samsum, rohde2021hierarchical}. However, unlike the short dialogues which contains less than 20 utterances, some tasks for summarizing much longer dialogues have been proposed recently \cite{chen2021summscreen, zhong2021qmsum}. These datasets are usually derived from meetings and interviews, with hundreds of turns in each dialogue. The length of such dialogues typically exceeds the input limits imposed by recent transformer-based models \cite{lewis-etal-2020-bart}, making it difficult to train an end-to-end summarization model for such tasks. This poses the challenge: \textit{How can we effectively use the current neural summarization models on dialogues that greatly exceed their length limits?}

Additionally, compared with document summarization, dialogues are interactive in nature, makes it more context-dependent and the information in dialogues is more sparsely distributed. Besides, the informal language used in dialogues leads to difficulties in modeling relevance and salience. To solve these issues, hierarchical methods are proposed to model the dialogues at turn level \cite{zhu2020a, rohde2021hierarchical}. However, generating a short summary that contains all the salient information remains challenging.

In this paper, we systematically investigate these issues on dialog summarization: we first explore the various solutions to the lengthy input problem. Then, we analyze and compare the methods to improve generic summarization models on challenging dialogue datasets. To address the long input issue, we investigate extended transformer models such as Longformer~\cite{beltagy2020longformer}, and several dialogue utterance retrieval methods for a \textit{retrieve-then-summarize} pipeline model, as well as hierarchical dialogue encoding models. 
For the specific challenges in dialogues, we explore different datasets for pretraining to test the transferability between similar summarization tasks. 
We evaluate these models on three recent long dialogue summarization datasets: QMSum for meetings~\cite{zhong2021qmsum}, MediaSum for interviews~\cite{zhu2021mediasum}, SummScreen for TV series transcripts~\cite{chen2021summscreen}.
In our experiments, we find that the pipeline method with a dialogue utterance retrieval model yields the best performance, and it can be further improved with a stronger retrieval model. Our experiment results also suggest that pretraining on proper external summarization datasets can effectively improve the performance of dialogue summarization models.
\section{Related Work}
\paragraph{Long Sequence Summarization}
\label{sec:long-summ-related-work}
Recent summarization models are based on Transformer \citep{vaswani2017attention} that has a quadratic time and memory complexity with respect to the input length, preventing it from being used for longer sequences. To address this issue, \citet{beltagy2020longformer} used the sliding window and global attention, while \citet{zaheer2020big} used a combination of random, sliding window and global attention mechanism to reduce the quadratic complexity to close-linear. 
Previous benchmarks for long sequence summarization mostly focus on documents instead of dialogues: \textsc{PubMed} and \textsc{Arxiv}~\cite{cohan2018discourse} consists of scientific papers which are typically very long; \textsc{BillSum}~\citep{kornilova-eidelman-2019-billsum} is a corpus of U.S. Congressional bills and their summaries; \textsc{BigPatent}~\citep{sharma-etal-2019-bigpatent} contains 1.3 million U.S. patent files and human-written summaries.

\paragraph{Dialogue Summarization}
Dialogue summarization aims to generate concise summaries for dialogues, such as meetings~\citep{mccowan2005ami, janin2003icsi, zhong2021qmsum,shang-etal-2018-unsupervised, zhu2020a}, TV series~\citep{chen2021summscreen}, interviews~\citep{zhu2021mediasum}, and chit-chat~\citep{gliwa-etal-2019-samsum, zhao-etal-2020-improving, chen2021structure}. Some summarization datasets (not limited to dialogues) contain queries asking for summarizing specific parts of dialogues~\citep{zhong2021qmsum, nema-etal-2017-diversity}, while others only need to summarize whole dialogues~\citep{chen2021summscreen, gliwa-etal-2019-samsum, DBLP:conf/nips/HermannKGEKSB15}. As for dialogue summarization models, \citet{zhu-etal-2020-hierarchical} described a hierarchical model for both inner- and cross-utterance attention, while \citet{chen-yang-2020-multi} proposed a multi-view decoder to leverage different extracted views of dialogues, such as topic view and stage view. 
\an{added the last part here}



\begin{table}[!ht]
\centering
\small
\resizebox{\linewidth}{!}{
\begin{tabular}{@{}lcccccc@{}}
\toprule
 & QMSum   & SummScreen & MediaSum*   \\ \midrule
Source                                          & Meeting & TV Series  & Interviews \\
Query-based                                     & YES     & NO         & NO         \\
\# examples                                      & 1.8k    & 26.9k      & 463.6k     \\
\# input tokens                            & 9069.8  & 6612.5     & 1553.7     \\
\# summary tokens                               & 69.6    & 337.4      & 14.4       \\
\# speakers                                     & 9.2     & 28.3       & 6.5        \\ \bottomrule
\end{tabular}
}
\caption{Comparison between three long dialogue summarization datasets we mainly study in this work. Numbers in the table are averaged across all samples. (*: MediaSum is only used for pretraining)}
\label{tab:dataset}
\end{table}

\section{Methodology}
In this section, we will introduce the dataset used to evaluate and pretrain the model, two types of summary models, and the details of the experiment setup.  
\an{change it to methodology?}
\subsection{Datasets}
To explore the problems in long dialogue summarization, we leverage three different long dialogue summarization tasks as main datasets: \\
\noindent\textbf{QMSum}~\citep{zhong2021qmsum} is a query-based multi-domain meeting summarization dataset annotated by humans. It contains 1,808 queries together with 232 long meeting transcripts, with topics as software product, academics, and committee meetings.\an{committe?} 
QMSum also contains annotated gold spans which could be used as the gold labels for training the retrievers; \\
\noindent\textbf{MediaSum}~\citep{zhu2021mediasum} is a large-scale media interview dataset consisting of 463.6K transcripts collected from NPR and CNN. Because MediaSum contains short summaries, i.e. only a short sentence representing the topic, we only use this dataset for pretraining and analysis. Due to the huge size of this dataset, 20k samples are randomly extracted for pretraining; \an{I don't think you can say "not naturally-written" unless it's machine generated, for which I don't think is the case.}\\
\noindent\textbf{SummScreen}~\citep{chen2021summscreen} is a dialogue summarization dataset consisting of 26.9k pairs of TV series transcripts and human-annotated summaries. 
It comes with two sources for recaps, and in this work, we choose one of them,i.e. ``Forever Dreaming'', for which we call SummScreen-FD as our benchmark.
\an{Consider changing it to "It comes with two sources for recaps, and in this work, we choose one of them, for which we call SummScreen-FD as our benchmark."}

\autoref{tab:dataset} shows the statistics for these three long dialogue datasets.
Additionally, we also consider \textbf{CNN/Dailymail}~\citep{DBLP:conf/nips/HermannKGEKSB15} (CNN/DM), \textbf{XSum}~\citep{xsum-emnlp}, and \textbf{SAMSum}~\citep{gliwa-etal-2019-samsum} as datasets for pretraining in our experiments.

\subsection{Models}
\subsubsection{Retrieve-then-summarize Pipeline}
\label{sec:utt-retrieval}
Dialogues tend to be relatively long, and most existing summarization models cannot process such long inputs.
The two-stage \textit{retrieve-then-summarize} pipeline first retrieves the most relevant subtext in the dialogue and then feeds to a summarizer.
We experiment with the following retrievers:
\begin{itemize}
    \item \textbf{TF-IDF}~\citep{jones1972statistical} Based on bag-of-words representation, TF-IDF measuers term frequency (TF) and normalizes them with inversed document frequency (IDF);
    \item \textbf{BM25}~\citep{robertson2009probabilistic} Similar to TF-IDF but accounts for document length and term saturation;
    \item \textbf{Locator}~\footnote{We obtained the locator output from the original authors.} The utterance locator model proposed by \citet{zhong2021qmsum} using convolution neural networks with BERT \citep{devlin2019bert}.
\end{itemize}

For TF-IDF and BM25, we limit the number of retrieved utterances to be at most 10\% of the whole dialogue, while we directly use the utterances predictor by Locator in its setting. After retrieval, we use the BART-large model fine-tuned on the output of the various retrievers to produce the summary.

\subsubsection{End-to-end Summarization Models}
To study how current state-of-the-art neural summarizers perform on long dialogue summarization, we choose the following three models:

\noindent\textbf{BART}~\citep{lewis-etal-2020-bart} is a transformer-based encoder-decoder model which obtains a number of state-of-the-art results on various text generation tasks. We use this model as our baseline summarization model for studying its ablations under different settings. The maximum number of input tokens is 1,024 so we truncate the input when it exceeds such limit.\footnote{We also tried to extend the positional embeddings to 2,048 for BART to accept longer input but found the results to be worse in our case.}

\noindent\textbf{HMNet}~\citep{zhu2020a} is a hierarchical network for dialogue summarization. It models the structure of the dialogue, using a token level encoder to encode each sentence and a turn level encoder for aggregating each turn. 
We use HMNet as a representative for the hierarchical type of models and compare it with other baselines. 
Due to the limitation of the memory cost, we constrain the maximum number of tokens to be 8,192 for HMNet, which is 8x as large as BART mentioned above.

\noindent\textbf{Longformer}~\citep{beltagy2020longformer} adapts the self-attention mechanism from full attention matrix to sliding window attention + global attention, which is more memory efficient. Longformer can accept up to 16K tokens and has shown improvement over long document summarization using its long-encoder-decoder (LED) variant. We allow the maximum input of 4,096 tokens for Longformer and cutoff the rest of the input, as we found further increasing such limit yields no improvements.

To incorporate queries in QMSum for these end-to-end models, we simply append the queries to the front of the meeting transcripts, as it is a standard practice for query-based summarization and also question answering \citep{devlin2019bert}\an{citation needed here}.

\subsection{Experiment Setup}
For a fair comparison between all models, we fit all of the models into the same RTX 8000 GPU with 48 GiB of GPU memory. We adopt the fairseq\footnote{https://github.com/pytorch/fairseq} implementation for BART, and the original code base for both Longformer\footnote{https://github.com/allenai/longformer} and HMNet\footnote{https://github.com/microsoft/HMNet}. 
We inherit the hyperparameters for all those models for fine-tuning in our experiments.\footnote{For more implementation details, please refer to our experiment code: https://github.com/chatc/LongDialSumm.}
Our most expensive experiments are fine-tuning for HMNet and Longformer, which take around 8 hours, while the runtime for BART model is less than one hour. We use ROUGE \cite{lin-2004-rouge} as our main evaluation metric and \texttt{pyrouge} library\footnote{https://github.com/bheinzerling/pyrouge} as the ROUGE implementation throughout all experiments.



\section{Result and Analysis}
Here we demonstrate our findings in four corresponding subsections. We also show some concrete examples and perform qualitative analysis in \autoref{sec:case-study}

\subsection{Dealing with Long Dialogues}
We compare several methods for addressing the long input issue for dialogue summarization, including different utterance retrieval methods describe in \autoref{sec:utt-retrieval} for a retrieve-then-summarize framework, heuristics for shortening the dialogue as well as baseline methods to establish reasonable bounds. 
From \autoref{tab:long-input-study}, we can see that even in the query-based dialogue summarization with QMSum, randomly selecting utterances still presents a strong baseline.
Over different modeling choices, the retrieve-then-summarize framework generally works better than end-to-end learning with dialogue cutoff at maximum input length. 
We do not observe an advantage of using Longformer over the BART model. This raises the question on whether all utterances in the dialogue are needed to produce a good summary or irrelevant utterances would add more noise.
Moreover, we notice that all these methods present a non-trivial gap with the summarization performance on the gold span, which uses relevant utterances annotated by humans. 
This suggests that there is plenty room for improvement if a better utterance retrieval method is developed. 
\begin{table}[!t]
\small
    \centering
    \begin{tabular}{lccc}
    \toprule
     Methods  & R-1 & R-2 & R-L \\  
     \midrule
    \multicolumn{2}{l}{\textbf{Retrieve-then-summarize}} & \\
    Random & 31.1 & 7.9 & 20.9 \\
    \hdashline
    TF-IDF &  32.5 & 8.5 & 21.4 \\
    BM25 &  \textbf{32.9} & \textbf{9.0} & \textbf{22.0} \\
    Locator & 29.9 & 7.6 & 19.6 \\
    \hdashline
    Gold span & 36.6 & 14.0 & 25.5 \\
    \midrule
    \multicolumn{2}{l}{\textbf{End-to-end (Cutoff at max \# tokens)}} & \\
     BART-large(1024) & 32.6 & 8.7 & 21.6 \\
     Longformer-large(4096) & 31.6 & 7.8 & 20.5 \\
    \bottomrule
    \end{tabular}
    \caption{Comparison of different methods for addressing the length of the dialogues on QMSum. All "retrieve-then-summarize" pipelines use BART-Large as a backend for summarization. "Gold span" denotes the annotated relevant turns in QMSum.
    }
    \label{tab:long-input-study}
\end{table}
\begin{table*}[!t]
\centering
\small
\begin{tabular}{lcccccc}
\toprule
                 & \multicolumn{3}{c}{QMSum} & \multicolumn{3}{c}{SummScreen-FD}  \\ 
                 & ROUGE-1      & ROUGE-2     & ROUGE-L     & ROUGE-1     & ROUGE-2     & ROUGE-L    \\ \hline
BART-Large            &   36.56      &    14.05     &  25.54   &    27.12    &    4.88    &    16.82      \\ \hline
\quad + XSum     &   34.90     &   13.49     &  24.90     &   27.17     &   4.59     &    17.02       \\
\quad + MediaSum   &    34.23 & 13.06 &  25.21       &    27.73     &    5.03     &    17.09      \\
\quad + CNN/DM        &     \textbf{39.88}     &  \textbf{15.94}      &  \textbf{28.02}      &    \textbf{28.86}     &   \textbf{5.55}      &     17.39        \\  \hline
\quad + CNN/DM-SAMSum &   35.46   &  12.52     &    24.62    &   28.15  &   5.41   &    17.25        \\
\quad + CNN/DM-MediaSum  &   36.79   &  13.69  &  25.94     &  28.68   &  5.31     &  \textbf{17.42}   \\
         \bottomrule
\end{tabular}
\caption{The performance of BART-large models that are pretrained on various summarization datasets. 
}

\label{tab:transfer}
\end{table*}

\subsection{Robustness to Input Length}
\begin{figure}[!t]
    \centering
    \includegraphics[width=\linewidth]{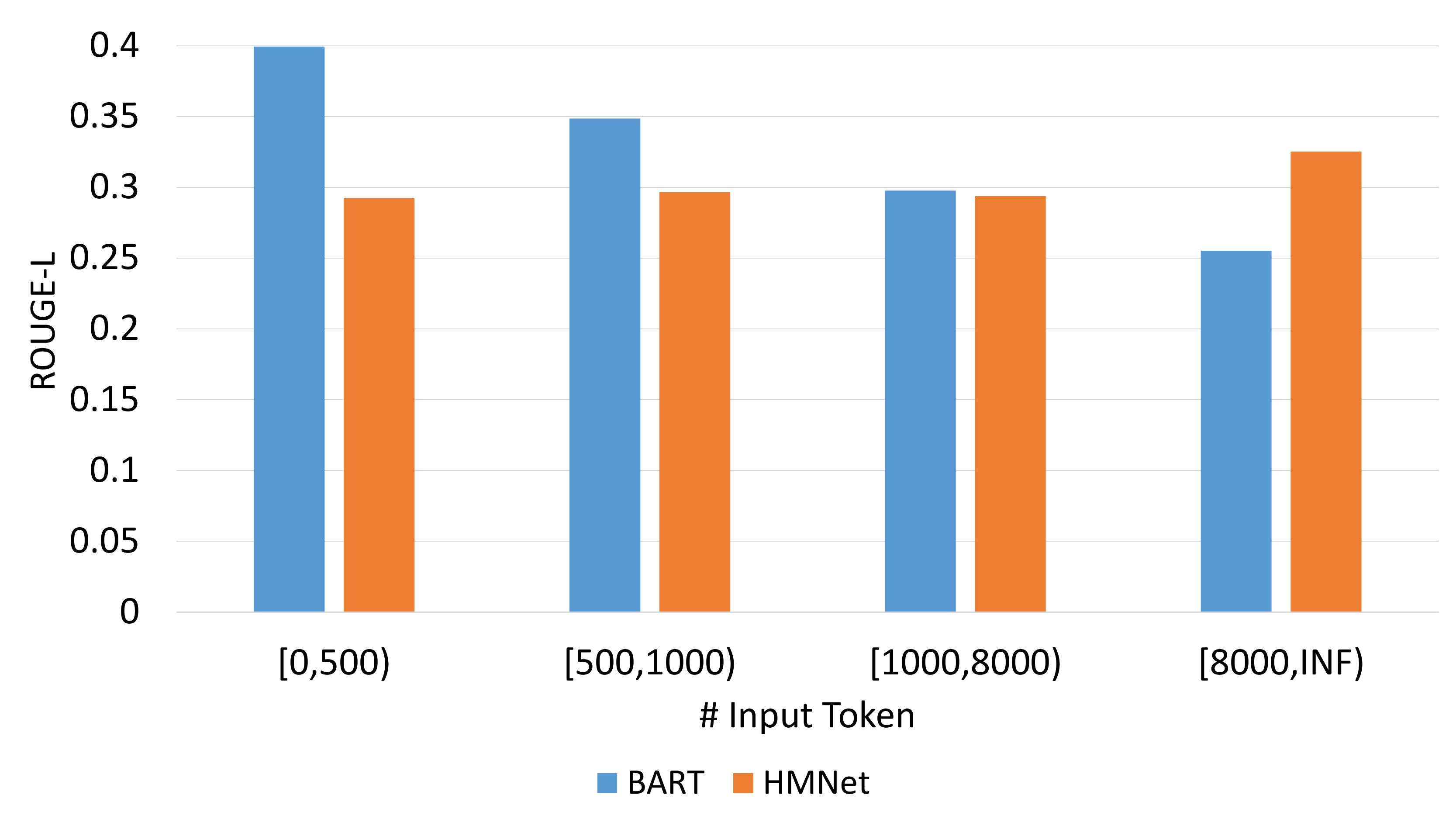}
    \caption{The ROUGE-L F1 scores of BART and HMNet on QMSum stratified by the number of input dialgue tokens.}
    \label{fig:input}
\end{figure}

\rui{How about we put this section as 4.2?}
As we discussed, some dialogues (e.g., QMSum) contain more than 20k tokens.
They exceed the input limitation of most existing summarization models. 
In this section, we further analyze the performance of summarization models as the input length changes.
To compare the robustness between two types of models (mainly BART and HMNet), we divide the test dialogues by the number of tokens. 
As we can see in \autoref{fig:input}, the performance of the BART model decreases sharply when the dialogue input becomes longer while the HMNet shows the opposite effect. 
This could be the result of their unique properties: BART is pretrained on the datasets with a limited length (\ie 1,024) and the input has to be truncated to fit the limitation, while HMNet obtains more information when the input is longer. However, the overall performance of HMNet is worst than BART. 

\begin{table}[!t]
\small
    \centering
    \begin{tabular}{lccc}
    \toprule
         & ROUGE-1 & ROUGE-2 & ROUGE-L \\  
    \midrule
    BART-CNN &  &  & \\
    w/o Query & 34.48 & 11.5 & 23.11 \\
    w/ Query & \textbf{39.88} & \textbf{15.94} & \textbf{28.02} \\
    \midrule
    HMNet & & & \\
     w/o Query & 35.1	& 10.1 & 30.8  \\
     w/ Query  & \textbf{36.8} & \textbf{10.9} & \textbf{31.9}\\
    \bottomrule
    
    \end{tabular}
    \caption{The performance comparison between BART and HMNet models on the query-based meeting summarization QMSum dataset. 
    }
    \label{tab:qmsum_query}
\end{table}

\subsection{Incorporating Queries}
Certain dialogue summarization tasks, such as QMSum, require generating a summary based on a specific question about the dialogue (e.g., opinion of a speaker or conclusion to a topic).
In this section, we study the influence of incorporating queries in dialogue summarization. 
\autoref{tab:qmsum_query} shows the performance of two models, BART and HMNet, on QMSum with and without queries at the beginning of the input. 
For the input to the two models, we use the gold relevant text spans given a query in QMSum to avoid the influences of retrieval models. The results show that encoding queries has a large impact on both types of models, especially for BART, even if the gold utterances are given. 

\subsection{Transfer Ability between Different Tasks}
\rui{How about put this section as 4.4?}
Pretraining has been shown effective for document summarization by introducing external knowledge from other similar tasks~\citep{DBLP:conf/nips/HermannKGEKSB15, fabbri-etal-2019-multi}.
We hypothesize that it is especially important for dialogue summarization because the dataset size is usually small.
Therefore, we study the transfer learning between different dialogue summarization tasks via pretraining.
\autoref{tab:transfer} shows the performance of BART-large models that are pretrained using different datasets and later fine-tuned on QMSum and SummScreen-FD. 
The results show that BART-large pretrained on CNN/Dailymail dataset (BART-CNN) yields the best performance after finetuning, though CNN/Dailymail consists of News articles and is not in dialogue format. We also note that pretraining on external datasets can also hurt the performance, and thus such pretraining datasets need to be carefully chosen. 

We also analyze the performance of BART-large by pretraining it on more than one dataset to test if BART-large can be further improved. We use the BART-large model pretrained on CNN/DM (BART-CNN) as baseline model since BART-CNN yields the best performance compared with the others. And then pretrain the same BART-CNN model on SAMSum and MediaSum separately. However, \autoref{tab:transfer} shows that after pretraining BART-CNN on these two datasets, ROUGE scores decrease sharply on QMSum dataset, and lightly on SummScreen-FD dataset except for ROUGE-L. This result demonstrates that pretraining on multiple dataset may not further improve the performance of the pretrained models.


\subsection{Case Study}
\label{sec:case-study}
We exam several summaries generated by BART-large model pretrained on three different datasets. We found that the BART-CNN model yields the best output with the least number of syntax errors and the closest content to the desired ones, while the output of BART-MediaSum model is usually shorter than Gold resulting in incomplete generation, and BART-XSum model usually predicts summaries with errors and duplication. This could be the result of data bias of pretraining datasets --- Summaries in MediaSum and XSum are shorter than CNN/DM. However, despite the better performance of BART-CNN model, these cut-off models fail to predict some part of the gold summary when the number of tokens in input dialogue is larger than the maximum input length of the model. For concrete examples, please refer to \autoref{sec:appendix}.

\section{Conclusion and Future Work}
We first explore the lengthy input problem of dialogue summarization through experiments on transformers and retrieval models. We conclude that the retrieve-summarize pipeline results in the best performance. Then, the experiments demonstrate the important role of queries and robustness to input length for different types of models. We found that adding a single query sentence in the input greatly improves ROUGE scores on QMSum. Additionally, BART performs worse when the input is beyond 512 tokens, even with extended positional embeddings; on the contrary, the hierarchical model performs better for longer inputs. We also test the transferability of summarization datasets by pretraining the language model on similar tasks. We conclude that the BART-large model pretrained on CNN/DM yields the best performance on both QMSum and SummScreen-FD.

For future work on solving the long input problem, we found that using an utterance retrieval model for summarization is a promising direction, yet modeling relevance between query and dialogue utterances remains a challenging task. And for the summarization models, it is worth exploring methods to 1) pretrain on valuable datasets for dialogue summarization, 2) better fuse the queries into the neural models, and 3) make the model robust to the input length (like HMNet) and maintain the high performance in the meantime (like BART).

\section*{Acknowledgments}
The authors would like to thank Ming Zhong, Da Yin, Yang Liu for their discussions and anonymous reviewers for their helpful comments. This work is supported in part by a grant from Microsoft Research.
\bibliography{anthology,custom}
\bibliographystyle{acl_natbib}

\appendix

\section{Generated Cases}
\label{sec:appendix}
\autoref{tab:case} shows some concrete sample summaries generated by BART-large model pretrained on various datasets.

\begin{table*}[!t]
\small
\resizebox{\textwidth}{!}{
\begin{tabular}{@{}lll@{}}
\toprule
         &  \makecell[c]{QMSum} & \makecell[c]{SummScreen-FD}\\ \midrule
         
MediaSum & \makecell[l]{
Transcribers are working on \green{transcribing} the data from the corpus . \\ The next step is to insure that the data are \green{clean} first , and then \\\green{channelized} . The transcribers are also working on ensuring\\ that the mark-up is consistent all the way throughout .} & \makecell[l]{
\green{Sydney} and \green{Will} are sent to a secret CIA project to find out \\if their father is alive or dead . Meanwhile , \green{Sydney} and\\ \green{Vaughn} are sent to a secret CIA facility to find out what\\ Irina is up to .
}\\\midrule

XSum     & \makecell[l]{
The transcribers have transcribed about thirty-five hours of \\ \green{transcripts} from the corpus . The next step is to insure that \\ the data is \green{clean} first , and then \green{channelized} . \red{The transcribers} \\\red{are working on is to insure that the data is clean first , and} \\ \red{then channelized} . The transcribers are also incorporating \\ additional conventions that Liz requested in terms of having a\\  systematic handling of numbers , acronyms and  acronyms which \\I had n't been specific about .} & \makecell[l]{
\green{Sydney} and \green{Will} are shocked to learn that Sydney's father , \\who was killed in Madagascar , is alive and working for the\\ CIA . \green{Will} is also shocked to learn that Sydney 's mother ,\\ who was killed in the Rambaldi experiment , is alive . Will\\ is also shocked to learn that Sydney 's father is a scientist . \\ \red{Will is also shocked to learn that Sydney 's mother is a}\\ \red{scientist} . \red{Will is also shocked to learn that Sydney 's} \\\red{mother is a scientist .} $\cdots$ \\ 
}\\ \midrule

CNN      & \makecell[l]{
The team was working on \green{transcribing} the data , and the next \\ step was to ensure that the data was \green{clean} first , and then \\ \green{channelized} . The team was working on ensuring that the data \\ was spell-checked , that the mark-up was consistent all the way \\ throughout , and that they incorporated additional \green{conventions} \\ that Liz requested in terms of having a systematic  handling \\of numbers , acronyms , and acronyms which they had  n't been \\specific about . }  & \makecell[l]{ 
\green{Sydney} and \green{Will} investigate the death of her father , who was\\ killed in a Russian KGB operation in 1982 . They discover that\\ the Rambaldi device was a Russian spy device , which was used \\to \green{test the IQ of children} . Sydney 's father was a KGB agent\\ , and she is now a KGB agent . She is also a double\\ agent , and she is working for the CIA . She is also working\\ for the CIA to find out who is behind the death of her father .\\ Meanwhile , Irina is worried about her father 's death ,\\ and she is worried about her relationship with \green{Vaughn} .
}\\ \midrule

Gold     & \makecell[l]{
Efforts by speaker fe008 are in progress to ensure that transcripts \\are clean ( i.e . spell checked ) , channelized , and conform \\to set conventions regarding the coding of numbers ,  acronyms , \\and explicit comments ( e.g . door slams , coughs , and laughter ) .\\ \gray{Subsequent efforts by speaker fe008 will be to tighten up boundaries} \\ \gray{on the time bins . Inter-annotator agreement was reported to be very }\\ \gray{good .Speaker mn014 's multi-channel speech/non-speech} \\ \gray{segmenter is in use .}} & \makecell[l]{ 
Sydney races to find a cure for Vaughn , but in order to find\\ the antidote , Sydney must make a deal with Sark that could\\ endanger Sloane 's life . Meanwhile , Will continues his\\ research for Vaughn and discovers some disturbing\\ inconsistencies involving 20-year - old standardized IQ tests .\\ Sydney finds out that Vaughn has a girlfriend . \\ }\\ \bottomrule
\end{tabular}
}
\caption{Sample output summaries of various pretrained models on QMSum and SummScreen. The summary $S$ of row $X$, column $Y$ indicates that BART-large model which is pretrained on $X$ dataset generates summary $S$ from test set of $Y$. The errors and duplication are marked in red. The out-of-boundary contents are marked in grey. Tokens marked in brown indicate the keywords emerged in Gold summary.} 
\label{tab:case}
\end{table*}


\end{document}